\documentclass[preprint,12pt]{elsarticle}
		
		\usepackage{amssymb}
		\usepackage{amsmath}
		
		\usepackage{amsfonts,amsthm}
		\usepackage{geometry}
		\usepackage{booktabs}
		\usepackage{float}
		\usepackage{multirow,subfigure,subcaption}
		\usepackage{hanging}
		\usepackage{nicematrix}
		\usepackage{rotating}
		
		\journal{}
		
\begin{document}
			
			\begin{frontmatter}
				
				
				
				\title{Alternate loss functions and regression models that achieve robustness to outliers by modulating the learning rate}
				
				\author[label1]{Mathew Mithra Noel\corref{cor1}}
				\ead{mathew.m@vit.ac.in}
				\author[label3]{Arindam Banerjee}
				\author[label2]{Yug D. Oswal}
				\author[label2]{Geraldine Bessie Amali D}
				\author[label1]{Venkataraman Muthiah-Nakarajan}
				
				\affiliation[label1]{organization={School of Electrical Engineering},
					addressline={Vellore Institute of Technology},
					city={Vellore},
					postcode={632 014},
					state={Tamil Nadu},
					country={India}}
				
				\affiliation[label3]{organization={Ernst and Young GDS},
					addressline={Godrej Waterside, Ring Rd, DP Block, Sector V, Bidhannagar},
					city={Kolkata},
					postcode={700 091},
					state={West Bengal},
					country={India}}
				
				\affiliation[label2]{organization={School of Computer Science and Engineering},
					addressline={Vellore Institute of Technology},
					city={Vellore},
					postcode={632 014},
					state={Tamil Nadu},
					country={India}}

				\cortext[cor1]{Corresponding Author}
				\begin{abstract}
					Most real-world datasets used for training supervised learning models are contaminated with noisy data and outliers leading to large prediction errors. This paper proposes a new approach for achieving robustness where the learning rate is modulated by a factor that is sensitive to outliers. In this approach a reduction of the learning rate is shown to be achieved by using alternate loss functions that are infinitely differentiable, strictly convex or quasiconvex and more closely approximate the absolute error than Huber and log-cosh losses. A comparison of the performance of regression models trained with different loss functions on a wide variety of benchmarks and datasets is presented to demonstrate the superior performance of the Square Root Loss (SRL) and Smooth Mean Absolute Error (SMAE) losses proposed in this paper. Two new robust linear regression models are presented. Highly vectorized robust parameter update formulae that take advantage of modern GPUs for both stochastic and batch gradient descent are presented.
					
				\end{abstract}
				
				
				
				
			\end{frontmatter}
			
				
				
				\section{Introduction}
				
			Artificial Neural Networks (ANNs) are a class of universal function approximators that learn a parametrized approximation to the target function through a Gradient Descent (GD) based optimization process \cite{goodfellow} \cite{jano} \cite{zhao}. Learning in ANNs is reduced to the problem of learning a finite set of real parameters, namely the weights and biases in the ANN model. Good parameters that result in a good approximation to the target function are computed by minimizing a loss (also known as cost) function that provides a measure of the difference between ANN outputs and targets \cite{amari}. An important aspect of the loss function is that it distills the performance of an ANN model over a dataset into a single scalar value. The loss function is a single, continuously differentiable, and usually convex real-valued function of all the parameters of the ANN model \cite{terven} \cite{tian} \cite{wang}. 
			
			Most real-world datasets are contaminated with outliers and noisy data. MAE is most robust to outliers but suffers from being non-differentiable at the origin. MAE also has large derivative values close to its minimum leading to instability and oscillations during training.Thus differentiable approximations to MAE namely Huber and Log-Cosh losses were introduced for robust regression tasks.
			
			Historically, only strictly convex loss functions have been considered since they possess at most one global minimum. Loss functions that are convex but not strictly convex can have multiple global minima that share the exact same value \cite{boyd2004convex}, \cite{rockafellar1970convex}. Convexity (Eq.\ref{Convexity}), in particular strict convexity significantly reduces the computational cost of optimization \cite{nesterov2003introductory}, \cite{bertsekas2003convex}. The problem of training an ANN is not a convex optimization problem in terms of the learnable parameters (weights and biases) even when a convex loss function is used because the output of an ANN is a highly non-convex function in general. However, convex loss functions are still very desirable since they do not introduce additional local minima. Eq.\ref{Convexity} defines a convex function and a similar definition with strict inequality is used for strictly convex functions.

				\begin{equation}
					f(\lambda x + (1 - \lambda)y) \leq \lambda f(x) + (1 - \lambda)f(y)
					\label{Convexity}
				\end{equation}
				$\quad \forall x, y \in \text{dom } f \text{ and } \lambda \in [0 , 1].$
				
				In this paper, we also consider strictly quasiconvex (\ref{Quasiconvex}) loss functions since this category of functions can also have at most one global minimum. Since quasiconvex functions share many of the desirable properties of convex functions, we propose a new quasiconvex loss function that more closely approximates the MAE loss than popular losses used for robust regression.

				\begin{equation}
					f(\lambda x + (1 - \lambda)y) < \max\left(f(x),f(y)\right)
					\label{Quasiconvex}
				\end{equation}
				$\quad \forall x, y \in \text{dom } f \text{ and } \lambda \in (0 , 1).$
				
			An equivalent definition to (\ref{Quasiconvex}) states that a function is quasiconvex iff all its sub-level sets are convex. Sublevel sets of a convex function are always convex, but the converse is not necessarily true. This motivates the generalization of convex functions and definition of quasiconvex functions in terms of level sets (\ref{sublevel sets}).
				
				A function $f: \mathbb{R}^n \to \mathbb{R}$ is quasiconvex iff its $\alpha$-sublevel set:
				\begin{equation}
					S_\alpha = \{x \in \text{dom } f \mid f(x) \leq \alpha\}
					\label{sublevel sets}
				\end{equation}
				is a convex set for every $\alpha \in \mathbb{R}$.
				
			Strictly quasiconvex functions share many of the advantages of convex functions. In particular, any local search algorithm like GD is guaranteed to find a global minimum avoiding problems associated with multiple local minima.
			
			It is also desirable to choose a loss function inspired by statistical estimation theory to compute the most probable parameters given the dataset. Historically, the Mean Square Error (MSE) and cross-entropy from Maximum Likelihood Estimation theory (MLE) loss functions have almost been universally used to train ANNs \cite{hui}. MLE theory assigns unknown parameter values that maximize the probability of observing the experimental data. The logarithm of the probability of observing the data is often used for convenience and results in the cross-entropy loss. Thus, MLE estimation proceeds by maximizing the logarithm of the probability of the observed data as a function of the unknown parameters. For convenience and historical reasons, the negative logarithm of the probability is minimized using GD in practice. 
			
			The MSE loss is most frequently used in regression analysis where continuous real values have to be predicted and Cross-entropy is used for classification problems where the targets are one-hot encoded vectors \cite{hastie}. Both MSE and cross-entropy loss functions are derived from MLE theory as already described. MSE is a maximum-likelihood estimate when the model is linear, and the noise is Gaussian. However, it is frequently used in practice even when these assumptions cannot be justified. For classification problems, both MSE and cross-entropy losses can be used, but learning is generally faster with cross-entropy as the gradient is larger due to the log function in cross-entropy loss. The cross-entropy loss, also referred to as Binary Cross Entropy (BCE) loss, is inspired by information theory and measures the difference between two probability distributions \cite{brownlee}. 
				
			The MSE loss accords an inordinate importance to fitting outliers due to the squaring of the error, resulting in final trained models that fit outliers more than the actual data. Thus, alternate loss functions that penalize outliers less harshly are of interest for regression tasks with outliers in the training data. In the following, we introduce the mathematical formalism of loss functions and propose alternatives to the MSE loss that perform significantly better on benchmark problems. 
			
			The main contributions of this work are
			
			\begin{itemize}
				
				\item Two new loss functions for robust regression that outperform the popular Huber and Log-Cosh losses are proposed.
				
				\item  A robust linear regression model is proposed.
				
				\item An extensive empirical comparison of different loss functions on a variety of benchmarks is presented.
				
			\end{itemize}

			\section{New loss functions for robust regression}
			
			Robust regression addresses the issues arising from the presence of significant outliers in the training data that disproportionately affect the parameter estimates \cite{khan} \cite{malek} \cite{varin}. Effective approaches of dealing with outliers typically involve schemes to down-weight the influence of outliers \cite{jajo} \cite{meer}. For regression problems, the Mean Square Error (MSE) is most frequently used. However, the MSE is very sensitive to outliers since all errors are penalized as the square \cite{hodson} \cite{wilmott}. If the training data for regression comes from a distribution with a ``heavy tail," the presence of many outliers will lead to large prediction errors if MSE is used during training since MSE assigns more importance to fitting the model to outliers than to the actual data \cite{brass} \cite{chai} \cite{qi}.  The Mean Absolute Error (MAE) loss proportionally penalizes all errors, but MAE is not differentiable. Another problem with the MAE loss is that it has a large derivative ($ \pm 1 $) close to the origin, leading to oscillations about the minimum during gradient descent. 
			
			Thus, a differentiable approximation to the Mean Absolute Error (MAE) with small derivative values close to the origin to avoid instabilities during training is desired. Differentiable alternatives to the MSE loss like the Huber loss \cite{huber64} \cite{huber92} have been explored in the past. This paper proposes two computationally cheap, infinitely differentiable smooth approximations to the MAE loss that approximate MAE more closely than the popular Huber and Log-Cosh losses. An approximation to MAE is derived in (\ref{approx1}). In this paper $e$ will be defined to be $e := y - \hat{y}$.

				\begin{align} \label{approx1}
					\begin{split}
						|e| &= max(-e,e)\\ 
						&= e*sgn(e)\\
						&\approx e*\tanh(\frac{e}{\alpha})\\
					\end{split}
				\end{align}
				
					In the above,  $\alpha$ can be chosen to make approximation as close as necessary. In the following, $\alpha$ is taken to be 1 or 2 for convenience. Another approximation to MAE is derived in (\ref{approx2}). 
				
				If $\epsilon$ is a small number:
				\begin{align} \label{approx2}
					\begin{split}
						|e| &= \sqrt{e^2}\\ 
						&\approx \sqrt{e^2 + \epsilon }\\
					\end{split}
				\end{align}
				
				Based on the above approximations (\ref{approx1}) and (\ref{approx2}), the Smooth Mean Absolute Error (SMAE) and Square Root Loss (SRL) are defined below to smoothly and closely approximate MAE. 
				
				\begin{equation}\label{SMAE}
					SMAE(e) = \frac{1}{N}\sum_{i=1}^N e_i\cdot\tanh(e_i)
				\end{equation}	
				
				\begin{equation}\label{SRL}
					SRL (e) = \frac{1}{N}\sum_{i=1}^N \sqrt{e_i^2 + \epsilon }
				\end{equation}

				The SMAE loss proposed in this paper has the following nice properties:
				
				\begin{enumerate}
					\item For small error values ($|e| << 1$), SMAE approximates the MSE, since $SMAE(e) = e\cdot\tanh(e) \approx e^2$.
					\item 	For large error values ($|e| >> 1$), $\tanh(e) \approx 1$ for $e>0$ and  $\tanh(e) \approx -1$ for $e<0$. So for large values of $e$, the SMAE loss approximates the MAE loss: $SMAE(e) = e\cdot\tanh(e) \approx |e|$.		
				\end{enumerate}
				
				For convenience, the definitions of different loss functions used in this paper are collected in Table \ref{reg} below:

				\begin{table}[H]					
					\centering
					\caption{Loss functions used for regression}
					\begin{tabular}{l|l}
						\hline
						\\ \textbf{Loss} & \textbf{Definition} \\  \hline
						MAE          &  $\frac{1}{N}{\displaystyle\sum_{i=1}^N} |e_i|$           \\ 
						MSE           & $\frac{1}{2N}{\displaystyle\sum_{i=1}^N} e_i^2 $           \\ 
						Huber         &  $\frac{1}{2N}{\displaystyle\sum_{i=1}^N} h((e_i) $ Where $h(e) =\frac{e^2}{2}$ for $|e|\le \delta$ and    $h(e)=\delta(|e| - \frac{\delta}{2})$ otherwise                \\ \\
						Log-Cosh      &  $\frac{1}{N}{\displaystyle\sum_{i=1}^N} \log(\cosh(e_i))$  \\\\ 
						SMAE          &  $\frac{1}{N}{\displaystyle\sum_{i=1}^N}  e_i\cdot\tanh(\frac{e_i}{\alpha})$           \\ \\
						SRL           &  $\frac{1}{N}{\displaystyle\sum_{i=1}^N} \sqrt{e_i^2 + \epsilon }$  \\ \hline
					\end{tabular}
					\label{reg}
				\end{table}
				
				\begin{figure}[h]
					\centering
					\caption{Plot of different loss functions used for regression.}
					\includegraphics[scale=0.25]{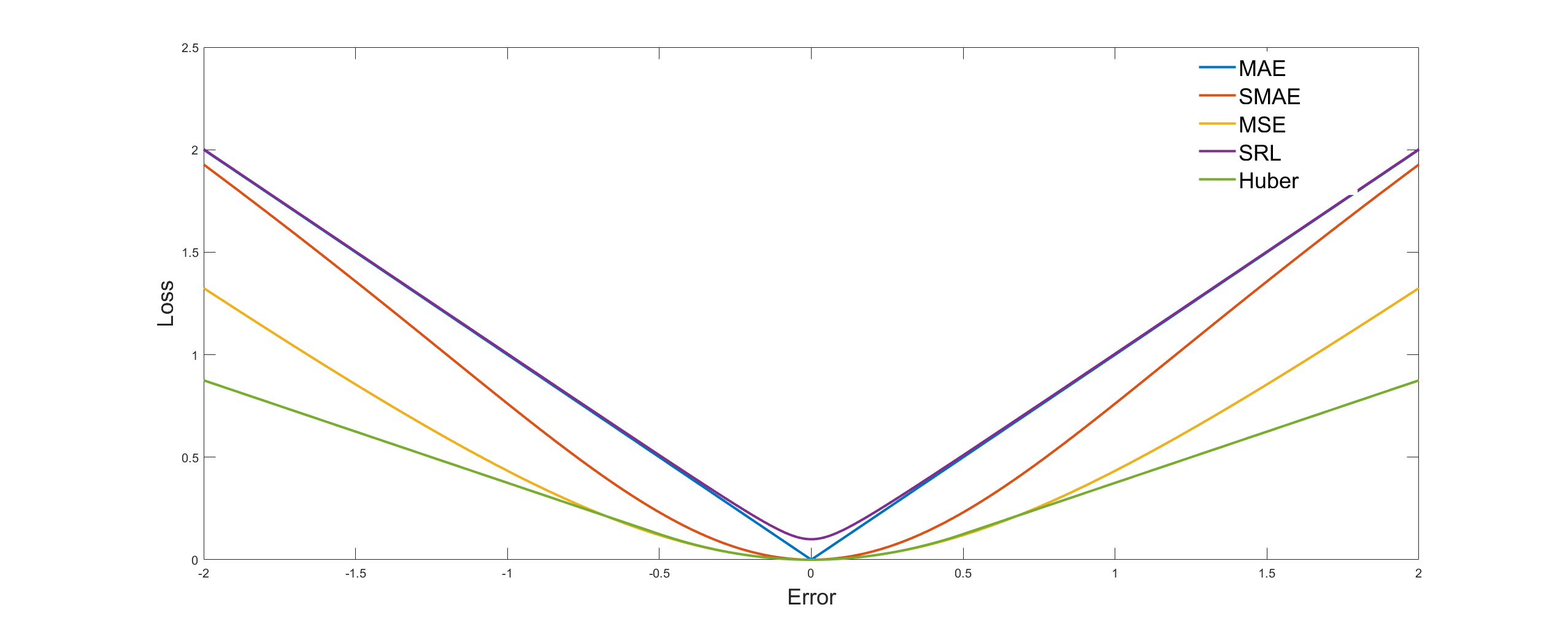}						
					\label{reg_losses}
				\end{figure}

			Figure. \ref{reg_losses} compares the SMAE and SRL losses proposed in this paper with different loss functions commonly used for regression.  The parameter $\delta$ in the Huber loss is taken to be 0.5, the parameter $\alpha$ in the SMAE loss is taken to be 1, and the parameter $\epsilon$ in the SRL loss is taken to be 0.01 in Figure. \ref{reg_losses} for clear visualization in the plots. It is clear from Figure. \ref{reg_losses} that SRL and SMAE approximate the ideal MAE loss more accurately than other loss functions used for regression. Also, it is clear from Figure. \ref{reg_losses} that SRL and SMAE losses converge to the ideal MAE loss for large error values, unlike the Huber \cite{huber92} and Log-Cosh losses \cite{saleh} that have a permanent offset with respect to the ideal MAE loss. Further, the parameter $\alpha$ can be chosen to be large to make SMAE as close to MAE as desired. Similarly, the parameter $\epsilon$ can be chosen to be small to make SRL as close to MAE as desired. 
			
			Both the SRL and SMAE loss functions are positive for all values of the error and attain their unique global minimum at the origin. The first and second derivatives of SRL are presented in  Eq. \ref{DSRL} and Eq. \ref{D2SRL}, respectively.

\begin{equation}
	\frac{d}{de}SRL(e) = \frac{e}{\sqrt{e^2 + \epsilon}} = \frac{e}{l}
	\label{DSRL}
\end{equation}

\begin{equation}
	\frac{d^2}{de^2}SRL(e) = \frac{\epsilon}{(e^2 + \epsilon)^{3/2}} = \frac{\epsilon}{l^3}
	\label{D2SRL}
\end{equation}

Since the second derivative of SRL is always strictly positive, SRL is strictly convex for all values by the second-order test for convexity. The SMAE loss is strictly quasiconvex due to the convexity of the open sublevel sets. Thus, both the proposed SRL and SMAE losses do not introduce new local minima. 

Also, unlike the popular Huber loss that fails to be differentiable twice, both the SRL and SMAE losses are infinitely differentiable (smooth). Thus, SRL and SMAE losses can be used with second-order optimization methods like Newton's methods, Trust-Region, Natural Gradient Descent (Information Geometry) and Cubic Regularization methods that require the second derivative.

\subsection{A Robust Linear Regression Model}
In the following, a SGD update rule for estimating the parameters of a linear model \cite{james} \cite{montgomery} with a custom loss function that is insensitive to outliers is presented. The predicted output of a linear regression model is:

\[\hat{y}	= \mathbf{w}^T\mathbf{x}+b\]

\begin{eqnarray}
	\text{Target value}	y & \in & (-\infty,\infty) \nonumber \\
	\text{Input vector    }\mathbf{x}&\in&\mathbb{R}^{d\times1} \nonumber \\
	\text{Output    }	\hat{y} & \in & (-\infty,\infty)  \nonumber\\
	\text{Weight vector    }	\textbf{w} &\in&\mathbb{R}^{d\times1} \nonumber \\
	\text{Bias parameter    }	b &\in& \mathbb{R} \nonumber \\
\end{eqnarray}

The SGD parameter update rule for any parameter $\Theta$, namely $\textbf{w}$ and $b$, is

\begin{equation} \label{sgd1}
	\Theta\leftarrow\Theta - \eta\frac{\partial l}{\partial\Theta}
\end{equation}

\begin{align} \label{chain}
	\begin{split}
		\frac{\partial l}{\partial b} &= \frac{\partial l}{\partial \hat{y}}.\frac{\partial \hat{y}}{\partial b}  \\
		\frac{\partial l}{\partial w_i} &= \frac{\partial l}{\partial \hat{y}}.\frac{\partial \hat{y}}{\partial w_i}  
	\end{split}
\end{align}

\begin{equation} \label{w}
	\frac{\partial \hat{y}}{\partial b} = 1 \\
	\frac{\partial \hat{y}}{\partial w_i} = x_i
\end{equation}	

\begin{equation}  \nonumber
	b\leftarrow b-\eta \frac{\partial l}{\partial b}= b - \eta\frac{\partial l}{\partial \hat{y}}.\frac{\partial \hat{y}}{\partial b} = b - \eta\frac{\partial l}{\partial \hat{y}} 
\end{equation}

\begin{equation} \nonumber
	w_i\leftarrow w_i - \eta \frac{\partial l}{\partial w_i}= w_i - \eta\frac{\partial l}{\partial \hat{y}}.\frac{\partial \hat{y}}{\partial w_i} = b - \eta\frac{\partial l}{\partial \hat{y}}x_i	
\end{equation}

\begin{equation}\label{SGD}
	\textbf{w}\leftarrow \textbf{w} - \eta \frac{\partial l}{\partial \hat{y}} \textbf{x} 	
\end{equation}

The derivative of the loss with respect to the error is presented in Table \ref{LossFunctionsList}. Since $l(e) = l = y - \hat{y}$:

\begin{equation}
	\frac{\partial l}{\partial \hat{y}} =  \frac{\partial l}{\partial e}\frac{\partial e}{\partial \hat{y}} = -\frac{\partial l}{\partial e} = -\frac{dl}{de}
\end{equation}

Substituting the expression for the derivative in Table \ref{LossFunctionsList}, the SGD update for robust regression with the proposed SRL loss is:

\begin{equation}\label{SRL update}
	\textbf{w}\leftarrow \textbf{w} + \eta\frac{e}{l}\textbf{x} 	
\end{equation}

Similarly, the SGD update formula for robust regression with the proposed SMAE loss is:

\begin{equation}\label{SMAE update}
	\textbf{w}\leftarrow \textbf{w} + \eta \left(\frac{l(1-l)}{e}+e \right)\textbf{x} 	
\end{equation}

Eq. \ref{SRL update} and \ref{SMAE update} must be compared with the standard SGD linear regression update formula Eq. \ref{linear regression}: 
\begin{equation}
	\textbf{w}\leftarrow \textbf{w} + \eta e \textbf{x} 
	\label{linear regression}
\end{equation}

Inspection of Eq. \ref{SRL update} and Eq. \ref{linear regression} reveals the $\frac{e}{l}$ factor that scales the learning rate. This $\frac{e}{l}$ factor reduces the magnitude of the update for very large loss values and prevents the model from selectively fitting outliers, thereby improving robustness. Similarly, the factor $\frac{l(1-l)}{e}$ is negative for large values of the loss, preventing large updates when outliers are encountered.

The above update equations implement an online or SGD update, for mini-batch/batch update (Eq. \ref{SMAE batch update} and Eq. \ref{SRL batch update}) the update term must be averaged over the appropriate batch of size $M$:

					\begin{equation}\label{SRL batch update}
					\textbf{w}\leftarrow \textbf{w} +  \frac{\eta}{M}\sum_{i=1}^M\frac{e_i}{l_i}\textbf{x}^i 	
			     	\end{equation}

				\begin{equation}\label{SMAE batch update}
					\textbf{w}\leftarrow \textbf{w} +  \frac{\eta}{M}\sum_{i=1}^M\left(\frac{l_i(1-l_i)}{e_i} + e_i \right) \textbf{x}^i 	
				\end{equation}
				
				\begin{table}[h]					
					\caption{Loss functions used to train regression models.}
					\centering
					\setlength{\tabcolsep}{1pt}
					\small
					\begin{tabular}{llll} \hline
						\textbf{Loss}                                                              & \textbf{Definition} ($l$) & \textbf{Derivative} ($\frac{dl}{de}$) & \textbf{Properties} \\ \hline
						\textit{\begin{tabular}[c]{@{}l@{}}Squared Error\end{tabular}}                                                     &   $\frac{e^2}{2}$                  &  $e$                   &  Convex, $C^\infty$                   \\
						\textit{\begin{tabular}[c]{@{}l@{}}Absolute\\ Error\end{tabular}} &  $|e|$                   &   $sgn(e)$ for $e \ne0$                  &    Convex, $C^0$                 \\
						\textit{Huber}   &                                                             $\left\{\begin{tabular}{l}
							$\frac{e^2}{2}$  \mbox{ for }  $|e|\le \delta$\\ $\delta(|e| - \frac{\delta}{2})$ \mbox{ otherwise} \end{tabular} \right.$
						&  $\left\{\begin{tabular}{l}
							$e$ for $|e|\le \delta$ \\    $\delta\cdot sgn(e)$ otherwise  \end{tabular} \right.$                & \begin{tabular}{l}
								Convex, Differentiable \\
								once $C^1$
							\end{tabular} \\
						\textit{Log-Cosh}                                                          & $\log(\cosh(e))$                    &  $\tanh(e)$                   &  Convex, $C^\infty$                \\
						\textit{\begin{tabular}[c]{@{}l@{}}Smooth Absolute\\ Error\end{tabular}} &  $e\tanh(\frac{e}{\alpha})$                   &  $ \frac{l(1-l)}{e}) + e $                    &  Quasi-Convex, $C^\infty$                  \\
						\textit{\begin{tabular}[c]{@{}l@{}}Square Root\end{tabular}}                                                       & 
						$\sqrt{e^2 + \epsilon}$                    &   $\frac{e}{l}$                  &   Convex, $C^\infty$  \\ \hline         
					\end{tabular}
				\label{LossFunctionsList}
				\end{table}
				
				Highly vectorized and computationally efficient parameter update formulae that take advantage of modern GPUs to perform mini-batch updates can be derived by defining the following. 	Here $d$ is the number of features (dimensionality) and $M$ is the batch size.
				
				$\textbf{X}$ -  $M\times(d+1)$ data matrix (including an extra column of 1s for the bias).
				
			    $\textbf{w}$ - $(d+1)\times1$ vector of parameters.
				
				$\textbf{y}$ - $M\times1$ vector of actual target values for the mini-batch.
				
				$\textbf{e}$ - $M\times1$ vector of errors.
				
				$\textbf{l}$ - $M\times1$ vector of losses.

				Let:
				\[	\textbf{e} = \textbf{y} - \textbf{X}\textbf{w} \]
				
				Rewriting Eq. \ref{SRL batch update} in matrix notation by observing that the matrix-vector product produces a linear combination of the columns of the matrix scaled by the components of the vector: 
				
				\[\textbf{w}\leftarrow \textbf{w} + \frac{\eta}{M}\left(\textbf{X}^T(\textbf{e} \oslash\textbf{l})     \right)\]

			    In the above $\oslash$ is the standard notation for Hadamard (elementwise) vector division. 
				 
				\begin{equation}
					\textbf{w}\leftarrow \textbf{w} + \frac{\eta}{M} \textbf{X}^T((\textbf{y} - \textbf{X}\textbf{w}) \oslash\textbf{l}) 
					\label{vectorized batch SRL}      
				\end{equation}
				
				Similarly, the batch update with SMAE loss can be shown to be:		
				
					\begin{equation}
					\label{vectorized batch SMAE}
					\textbf{w}\leftarrow \textbf{w} + \frac{\eta}{M} \textbf{X}^T((\textbf{y} - \textbf{X}\textbf{w}) \odot (\frac{\textbf{l}(\textbf{1}-\textbf{l}) + \textbf{e}^2}{\textbf{e}}))     
				\end{equation}
				
				The vectorized mini-batch update that takes advantage of GPU hardware are given in Eq. \ref{vectorized batch SMAE} and Eq. \ref{vectorized batch SRL}. In the above, $\textbf{1}$ is the vector whose components are all 1s, $\odot$ is the standard notation for Hadamard (elementwise) multiplication of vectors and $\textbf{e}^2$ represents elementwise squaring.
				
\begin{figure}[!tbh]			        
			        \hspace{-3cm}
			    	\includegraphics[scale=0.4]{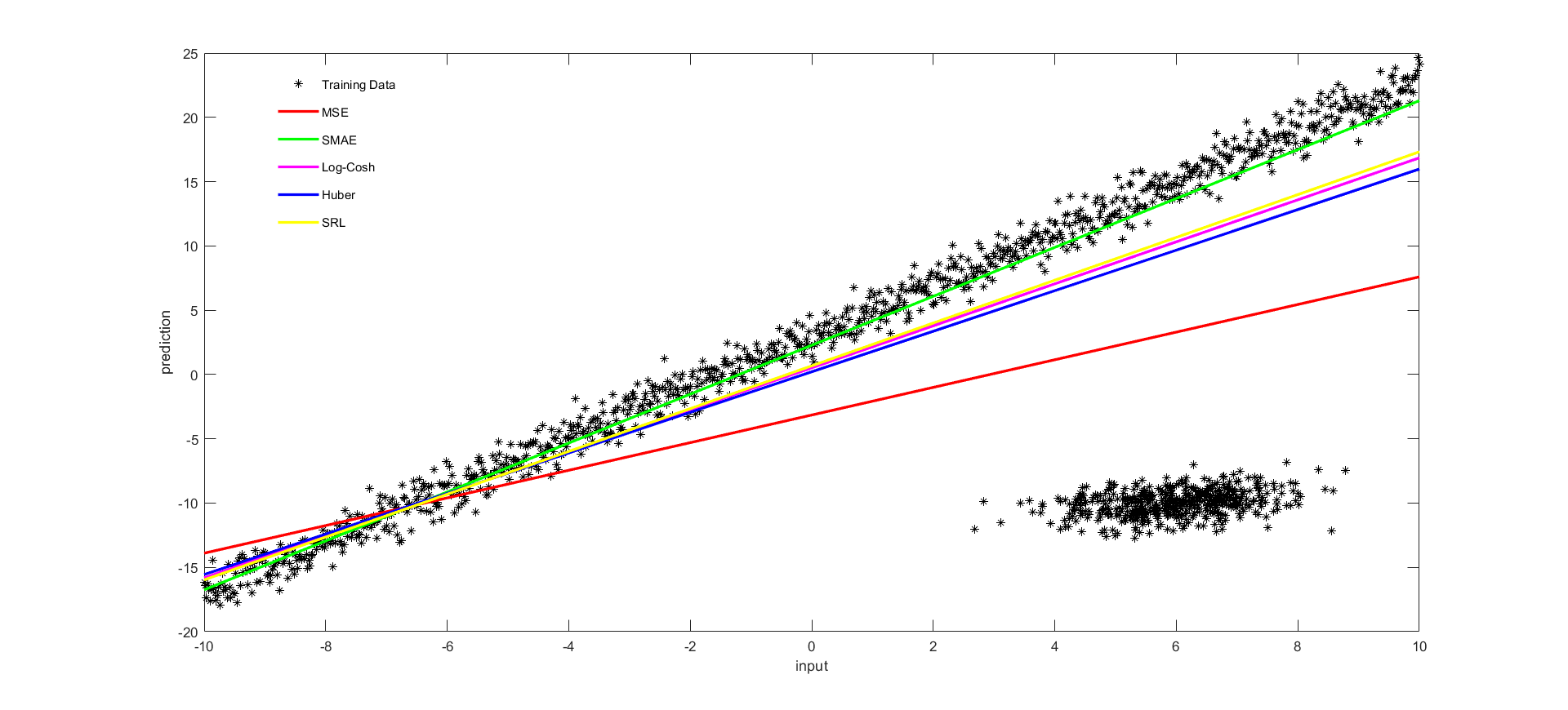}
					\caption{Effect of using different loss functions for linear regression when the dataset has a significant number of outliers. The training dataset is shown as black dots and consists of 1500 instances with 500 outliers.}
					\label{reg_plots}
				\end{figure}

				Using (\ref{SGD}), the performance of a linear regression model with different loss functions on a dataset with 33\% outliers is presented in the following. The training dataset consisted of 500 outliers and 1000 points (Figure. \ref{reg_plots}) generated by adding zero mean unit variance Gaussian noise to a linear model $y = 2*x+3$. A learning rate of 0.001 was used, and the slope as well as the Y-intercept were initialized to zero. The 500 outliers were generated from a 2D Gaussian distribution with mean vector $\mathbf{\mu}$ and covariance matrix $\mathbf{\Sigma}$ given in (\ref{mu}) and (\ref{sigma}).
			
				\begin{equation}  \label{mu}
					\mathbf{\mu} = 
					\begin{bmatrix}
						6\\
						-10\\
					\end{bmatrix}
				\end{equation}		
				
				\begin{equation} \label{sigma}
					\mathbf{\Sigma}=
					\begin{bmatrix}
						1 & 0.3\\
						0.3 & 1\\
					\end{bmatrix}
				\end{equation}

				\begin{table}[H]
					\caption{Parameter estimation error when different loss functions are used to fit a linear model to the dataset in Figure. \ref{reg_plots}.}
					\centering
					\begin{tabular}{c|c|c}\hline				    
						\textbf{Loss}     & \textbf{\begin{tabular}[c]{@{}l@{}}Slope\\ \% error\end{tabular}} & \textbf{\begin{tabular}[c]{@{}l@{}}Y Intercept\\ \% error\end{tabular}} \\ \hline
						\textit{MSE}      & 53.4                                                                 & 204.3                                                                      \\
						\textit{SMAE}     & 5.3                                                                 & 20.9                                                                       \\
						\textit{Log-Cosh} & 21.9                                                                & 83.6                                                                      \\
						\textit{Huber}    & 23.3                                                                & 93.5                                                                       \\
						\textit{SRL}      & 19.9                                                                 & 78.7     \\ \hline                                                                 
					\end{tabular}
					\label{table_regression}
				\end{table}

				Figure. \ref{reg_plots} shows that the SMAE and SRL losses proposed in this paper are less affected by outliers than other loss functions. The robustness of the proposed loss functions can be attributed to the closeness to the ideal MAE loss compared to other loss functions. Table \ref{table_regression} shows the percentage parameter estimation error when different loss functions are used to estimate the parameters of a linear model. In the following we provide an extensive comparison of the proposed loss functions for robust regression (SMAE and SRL) with the most popular loss functions on a variety of benchmarks and datasets.

				\begin{table}[!tbh]
					\centering
					\caption{Experimental setup for Synthetic Object Localization (MNIST)}
					\scriptsize
					\begin{tabular}{c|c} \hline
						Data Set & Synthetic MNIST \\ \hline
						Training Set&  60000 \\
						Validation Set & $5\%$ of Preprocessed Training Data \\
						Test Set & 10000  \\
						Model & CNN $\rightarrow$ Flatten $\rightarrow$ Feedforward Network \\ 
						Optimizer & Adam \\
						Batch Size & 32 \\
						Epochs & 20 \\ 
						Trials & 30  \\\hline
					\end{tabular}
					\label{mnist_setup}
				\end{table}
				
				\begin{table}[!tbh]
					\centering
					\scriptsize
					\caption{Various Loss Functions on MNIST Object Localization 30 Benchmark Datasets}
					\begin{tabular}{c|c|c|c|c|c} \hline
						\textbf{Loss Function} &	\textbf{Train RMS} & \textbf{Train MAE}& \textbf{Train R2}&\textbf{Test RMS}&\textbf{Test MAE}\\ \hline 
						MSE & 0.1447 & 0.1104 & 0.5859 & 0.0843 & 0.0715  \\
						SMAE & 0.1427 & 0.1089 & 0.6040 & 0.0815 & 0.0704   \\
						SRL & 0.1413 & 0.1026 & 0.6118 & 0.0705 & 0.0595  \\ 		   
						Log Cosh & 0.1434 & 0.1099 & 0.5997 & 0.0849 & 0.0719  \\
						Huber Loss & 0.1436 & 0.1115 & 0.5989 & 0.0875 & 0.0749  \\ \hline   
					\end{tabular}
					\label{mnist}
				\end{table}
		
	\subsection{Synthetic MNIST Object Localization Benchmark}
	Object localization is the problem of finding the spatial coordinates of objects in an image and has numerous applications to medical imaging, autonomous vehicles, surveillance, and augmented reality. Object localization is a regression problem, as it involves predicting the spatial coordinates of bounding boxes around objects. The Synthetic MNIST Object Localization benchmark is a variation of the popular MNIST classification benchmark and contains 60,000 images of handwritten digits in various positions in the images. This was done by placing the 28×28 MNIST digit randomly on a 75×75 canvas and generating a 28×28 bounding box around that digit. The top-left corner position of each digit's image is sampled from a uniform distribution with a minimum 0 and a maximum of 48. In order to evaluate the robustness of each loss function, 20\% outliers were synthetically added by randomly shifting the bounding box by 20–30 pixels to the bottom and right according to an uniform probability distribution. Table \ref{mnist_setup} provides the experimental setup, and Table \ref{mnist} compares the performance of different loss functions on the MNIST digit localization benchmark with 20\% outliers. Results presented in Table \ref{mnist} clearly indicate the superior performance of the proposed SMAE and SRL loss functions.

		\begin{table}[!tbh]
			\centering
			\caption{Experimental Setup for California Housing Dataset}
			\scriptsize
			\begin{tabular}{c|c} \hline
				Data Set & California Housing \\ \hline
				Training Set& 16512 preprocessed samples containing 8 input features with artificially introduced outliers \\
				Test Set & 4128 preprocessed samples with no outliers \\
				Validation Set & None \\
				Preprocessing & Min-max normalization, Add 0.5 noise to 20\% of the training labels \\		
				Model & Feedforward Network \\ 
				Optimizer & Adam \\
				Batch Size & 32 \\
				Epochs & 30 \\ 
				Trials & 50  \\ \hline
			\end{tabular}
			\label{Housing dataset}
		\end{table}
		
		\begin{table}[!tbh]
			\centering
			\scriptsize
			\caption{Various Loss Functions on California Housing Benchmark Datasets 50 trials}
			\begin{tabular}{c|c|c|c|c|c} \hline
				\textbf{Loss Function} &	\textbf{Train RMS} & \textbf{Train MAE}& \textbf{Train R2}&\textbf{Test RMS}&\textbf{Test MAE}\\ \hline 
				MSE & 0.2838 & 0.2316 & 0.0303 & 0.2088 & 0.1779 \\
				SMAE & 0.2839 & 0.2273 & 0.0297 & 0.2016 & 0.1696   \\ 
				SRL & 0.3017 & 0.2362 & -0.0899 & 0.2057 & 0.1691 \\
				Log Cosh & 0.2772 & 0.2241 & 0.0881 & 0.2002 & 0.1697  	\\
				Huber Loss & 0.2771 & 0.2242 & 0.0888 & 0.2004 & 0.1700 \\ \hline		
			\end{tabular}
			\label{cal_housing}
		\end{table}
		
\subsection{California House Price Prediction Benchmark}
This benchmark is widely used and is a standard reference for evaluating regression algorithms and contains median house values for districts in California from the 1990 U.S. Census \cite{fard}. This dataset contains 20,640 samples with features such as median income, housing age, total rooms, total bedrooms, population, households, and geographic location (longitude and latitude). The target (prediction) variable is the median house value. Table \ref{Housing dataset} shows the experimental setup, and Appendix I provides the ANN model used for this benchmark. Experimental results in Table \ref{cal_housing} show that the proposed SRL and SMAE losses help the model achieve higher test accuracies.

\subsection{Concrete Strength Prediction Benchmark}
The relationship between raw materials used and concrete strength is highly nonlinear, and learning this complex relationship is a challenging regression problem and hence frequently used to benchmark models \cite{slonski}. This dataset contains 1,030 samples and focuses on predicting compressive strength (in MPa) using eight input variables: cement, blast furnace slag, fly ash, water, superplasticizer, coarse aggregate, fine aggregate, and age of the concrete. Table \ref{concrete setup} shows the experimental setup, and Appendix I provides the ANN model used for this benchmark. Experimental results in Table \ref{concrete} show that the proposed SRL and SMAE losses achieve higher test accuracies compared to the most popular loss functions.

		\begin{table}[h]
			\centering
			\caption{Experimental Setup for Concrete Strength}
			\scriptsize
			\begin{tabular}{c|c} \hline
				Data Set & Concrete Strength \\ \hline
				Training Set& 824 preprocessed samples containing 8 input features with artificially introduced outliers \\
				Test Set & 206 preprocessed samples with no outliers   \\
				Validation Set & None \\
				Preprocessing & Min-max normalization, Add 0.5 noise to 30\% of the training labels \\
				Model & Feedforward Network \\ 
				Optimizer & Adam \\
				Batch Size & 16 \\
				Epochs & 50 \\ 
				Trials & 30  \\ \hline
			\end{tabular}
			\label{concrete setup}
		\end{table}
		\vspace{1cm}
		\begin{table}[h]
			\centering
			\scriptsize
			\caption{Various Loss Functions on Concrete Strength Benchmark Datasets 30 Trials}
			\begin{tabular}{c|c|c|c|c|c} \hline
				\textbf{Loss Function} &	\textbf{Train RMS} & \textbf{Train MAE}& \textbf{Train R2}&\textbf{Test RMS}&\textbf{Test MAE}\\ \hline 
				MSE & 0.3249 & 0.2823 & -0.3813 & 0.2424 & 0.2126  \\
				SMAE & 0.2996 & 0.2565 & -0.1402 & 0.2143 & 0.1858   \\
				SRL & 0.3143 & 0.2638 & -0.2718 & 0.2142 & 0.1842   \\ 
				Log Cosh & 0.2991 & 0.2575 & -0.1373 & 0.2175 & 0.1889  \\
				Huber Loss & 0.3641 & 0.3157 & -0.7495 & 0.2691 & 0.2379	\\ \hline		 
			\end{tabular}
			\label{concrete}
		\end{table}
		\vspace{1cm}
\subsection{Wine Quality Prediction Benchmark}
This benchmark is challenging because it provides an unbalanced dataset containing 1,599 samples for red wine and 4,898 samples for white wine \cite{cortez}. Each input sample is characterized by features such as fixed acidity, volatile acidity, citric acid, residual sugar, chlorides, free and total sulfur dioxide, density, pH, sulfates, and alcohol content. The target variable is the wine quality score, and this benchmark is an example of a highly unbalanced regression dataset. Table \ref{wine setup} shows the experimental setup, and Appendix I provides the ANN model used for this benchmark. Experimental results in Table \ref{winequality} show that the proposed SRL and SMAE losses again achieve higher test accuracies compared to the most popular loss functions.

		\begin{table}[h]
			\centering
			\caption{Experimental Setup for Wine Quality}
			\scriptsize
			\begin{tabular}{c|c} \hline
				Data Set & Wine Quality \\ \hline
				Training Set& 5197 preprocessed samples containing 11 input features with artificially introduced outliers  \\
				Test Set & 1300 preprocessed samples with no outliers  \\
				Validation Set & None \\
				Preprocessing & Min-max normalization, Add 0.5 noise to 20\% of the training labels \\
				Model & Feedforward Network \\ 
				Optimizer & Adam \\
				Batch Size & 32 \\
				Epochs & 50 \\ 
				Trials & 50  \\ \hline
			\end{tabular}
			\label{wine setup}
		\end{table}
		
		\begin{table}[h]
			\centering
			\scriptsize
			\caption{Various Loss Functions on Wine Quality Benchmark Datasets 50 Trials}
			\begin{tabular}{c|c|c|c|c|c} \hline
				\textbf{Loss Function} &	\textbf{Train RMS} & \textbf{Train MAE}& \textbf{Train R2}&\textbf{Test RMS}&\textbf{Test MAE}\\ \hline 
				MSE & 0.2943 & 0.2436 & -0.7438 & 0.2108 & 0.1824  \\
				SMAE & 0.2633 & 0.2104 & -0.3127 & 0.1788 & 0.1500   \\
				SRL & 0.2568 & 0.1982 & -0.2140 & 0.1619 & 0.1338    \\ 
				Log Cosh & 0.2710 & 0.2192 & -0.4201 & 0.1878 & 0.1591 \\
				Huber Loss & 0.2787 & 0.2287 & -0.5273 & 0.1985 & 0.1698 \\ \hline 				 
			\end{tabular}
			\label{winequality}
		\end{table}
		\vspace{2cm}
		\begin{table}[h]
			\caption{Specification of workstation used to train different models}
			\centering
			\begin{tabular}{l} \hline
				GPU: 2 $\times$ NVIDIA RTX A5000 24GB VRAM Each \\
				CPU: Intel (R) Xeon (R) Gold 6246R CPU at 3.40GHz \\ \hline
			\end{tabular}
			\label{computation}
		\end{table}
		
\section{Conclusion}

Loss functions are universally used in training ML models and different loss functions can result in models with widely differing final performances. This performance difference can be attributed to the critical role played by the importance assigned to small and large errors and differing derivative values leading to completely different trajectories during gradient-based training/optimization. This paper explored the possible advantages of using alternatives to popular loss functions for robust regression. Many real-world datasets are contaminated with significant numbers of large outliers, and these outliers can confuse regression algorithms, leading to large prediction errors. This paper proposed two infinitely differentiable, smooth, computationally cheap approximations to the Mean Absolute Error (MAE) loss referred to as the SMAE and SRL losses. In particular, the popular Huber loss does not converge to MAE for large errors because of restrictions imposed by convexity. This paper proposed the strictly quasiconvex SMAE loss that inherits the advantage of a unique global minimum while converging to MAE for large errors. The SRL loss proposed in this paper provides a very close strictly convex and smooth (infinitely differentiable) approximation to MAE. Extensive comparison on 5 benchmarks clearly indicates the superior performance of the proposed SMAE and SRL losses compared to popular loss functions like Huber and Log-Cosh used for datasets contaminated with outliers. 

Linear regression still remains the most widely used model in statistics and machine learning due to its low computational cost and interpretability despite the popularity of deep learning. This paper proposed two robust linear regression models (Eq. \ref{SMAE update} and Eq. \ref{SRL update}) that allow linear regression models to be trained on data heavily contaminated with outliers. The proposed robust linear regression models were shown to achieve robustness to outliers by reducing the learning rate for large errors. In particular, a factor of $\frac{e}{l}$  and a factor of $\left(\frac{l(1-l)}{e}+e \right)$ were shown to reduce the effective learning rate for large errors in the SRL and SMAE based robust linear regression models, respectively. Finally, vectorized mini-batch update equations that take advantage of modern GPU hardware were presented.

		\section*{Appendix I: List of ANN models}
		
		\begin{figure}[H]
			\centering
			\includegraphics[scale=0.35]{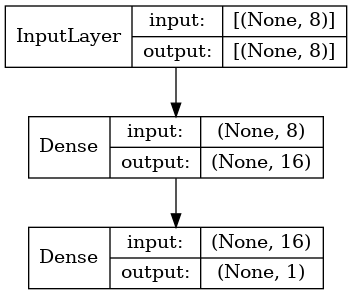}
			\caption{California housing 3D plot}
			\label{fig1}
		\end{figure}	
		
		\begin{figure}[!tbh]
			\centering
			\includegraphics[scale=0.35]{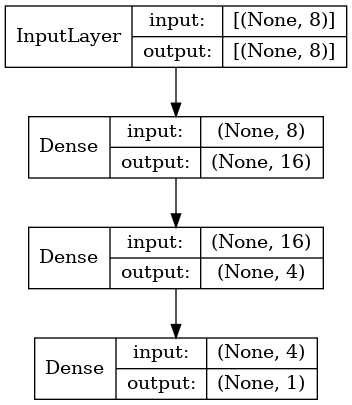}
			\caption{Concrete strength 3D plot}
			\label{fig2}
		\end{figure}	
		
		\begin{figure}[!tbh]
			\centering
			\includegraphics[scale=0.35]{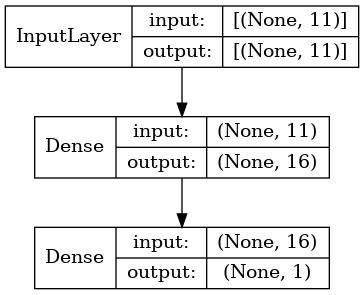}
			\caption{Wine quality 3D plot}
			\label{fig3}
		\end{figure}

		\end{document}